\documentclass[conference]{IEEEtran}
\IEEEoverridecommandlockouts
\usepackage{cite}
\usepackage[latin9]{inputenc}
\usepackage{amsmath}
\usepackage{graphicx}
\usepackage{multirow}
\usepackage{algorithm}
\usepackage{algorithmic}
\usepackage{array}
\usepackage{stfloats}
\usepackage{color}
\usepackage{amssymb}
\usepackage{float}

\ifCLASSINFOpdf
\else
\fi

\ifCLASSOPTIONcompsoc
  \usepackage[caption=false,font=normalsize,labelfont=sf,textfont=sf]{subfig}
\else
  \usepackage[caption=false,font=footnotesize]{subfig}
\fi

\def\BibTeX{{\rm B\kern-.05em{\sc i\kern-.025em b}\kern-.08em
    T\kern-.1667em\lower.7ex\hbox{E}\kern-.125emX}}
\begin{document}

\title{Improved Xception with   Dual Attention Mechanism and Feature Fusion for Face Forgery Detection}

\author{
\IEEEauthorblockN{ Hao Lin}
\IEEEauthorblockA{Guangdong Key Lab  of Information Security Technology\\
Sun Yat-sen University\\
Guangzhou, China \\
linh233@mail2.sysu.edu.cn}
\\
\IEEEauthorblockN{Kangkang Wei}
\IEEEauthorblockA{School of Computer Science and Engineering \\
Sun Yat-sen University\\
Guangzhou, China \\
 weikk5@mail2.sysu.edu.cn}

\and
\IEEEauthorblockN{Weiqi Luo}
\IEEEauthorblockA{School of Computer Science and Engineering \\
Sun Yat-sen University\\
Guangzhou, China \\
luoweiqi@mail.sysu.edu.cn}
\\
\IEEEauthorblockN{Minglin Liu}
\IEEEauthorblockA{School of Computer Science and Engineering \\
Sun Yat-sen University\\
Guangzhou, China \\
liumlin6@mail2.sysu.edu.cn}
}
\maketitle

\begin{abstract}
With the rapid development of deep learning technology,  more and more  face forgeries by deepfake are widely spread on social media, causing serious social concern.  Face forgery detection has become a research hotspot  in recent years, and  many related methods have been proposed until now.  For those images with low quality  and/or diverse sources,  however, the detection performances of existing methods are still far from satisfactory.  In this paper,  we propose an improved Xception with dual attention mechanism and feature fusion for  face forgery detection.  Different from the middle flow in original Xception model, we try to catch different high-semantic features of face images using different levels of convolution,  and introduce the  convolutional block attention module and feature fusion to refine and reorganize those high-semantic features.  In the exit flow, we employ the self-attention mechanism and  depthwise separable convolution to learn the global information and  local information of the fused features separately to improve the classification ability of the proposed model.   Experimental results evaluated on three Deepfake datasets demonstrate that the proposed  method  outperforms  Xception  as well as other related  methods both in  effectiveness and generalization ability. 
\end{abstract}


\IEEEpeerreviewmaketitle

\section{Introduction}
With the rapid development of deep learning technology,  various face forgery methods \cite{Thies_2016_CVPR,10.1145/3306346.3323035,karras2019stylebased,faceswap,deepfake} based on deep learning have been proposed.    Nowadays, more and more forged faces without any visual artifacts are widely spread on social media.  If these forgery images are abused, it probably leads to  a series of moral, ethical, and safety issues. Nowadays,  face forgery detection is facing severe challenges.


Up to now,  many deepfake detection methods \cite{mccloskey2018detecting,yu2018attributing, Huaxiao2018,guarnera2020deepfake,
9072088,li2018ictu,nguyen2019multitask,capsule,li2020face,mesoNet,zhao2021multiattentional} have been proposed.  
Most existing methods are mainly based on Convolutional Neural Network (CNN).  For instance,  in \cite{li2018ictu, 9072088}, the authors    make full use of physiological signal, such as the inconsistency of human blinks, to detect deepfake; 
in \cite{capsule}, the authors first introduce the capsule network for face forgery detection,  and show that  it can achieve very good detection performances compared with related methods;   
in \cite{yu2018attributing}, the authors try to detect those artifacts generated by  GAN (Generative adversarial network) for exposing face forgery;  in \cite{zhao2021multiattentional}, the authors consider deepfake detection as a fine-grained classification problem, and propose a new multi-attentional detection network. 
 Typically, the existing detection methods can achieve  good detection accuracy for high-quality databases.  However, the detection accuracies would drop  compared with their high-quality versions. Furthermore, for those face images with more diverse contents, such as  images from  WildDeepfake\cite{zi2021wilddeepfake},  the detection performances of existing methods are still far from satisfactory. 
 
The previous results in  \cite{rssler2019faceforensics} show that  Xception   \cite{chollet2017xception}  has better sensitivity to manipulated face images created by deepfake,  and it can outperform many related methods for face forgery detection.  In this paper, therefore, we use the Xception as the backbone of the proposed method,  and introduce  dual attention mechanism and feature fusion in the middle flow and exit flow in the original Xception model for face forgery detection.  In the middle flow, we firstly obtain different     high-dimensional features using different levels of convolution,  and  refine the features via Convolutional Block Attention Module (CBAM)  simultaneously.  Finally, we fuse them to  get a more comprehensive high-dimensional  features for subsequent network analysis. In the exit flow,  we employ the self-attention mechanism and  depthwise separable convolution to learn the global information and  local information of the fused features separately to improve the classification ability of the proposed model.  Experimental results show that the proposed method outperforms the related methods both in effectiveness and generalization ability. 

The rest of the paper is organized as follows. Section \ref{sec:Relative} describes related works. Section \ref{sec:Proposed} introduces the proposed method in detail. Section \ref{sec:experiments} shows comparative results and discussions. Finally, the concluding remarks of this paper and future works are given in Section \ref{sec:Conclusion}.

\section{Related work}
\label{sec:Relative}

In this section, we will describe three related works used in the proposed method, that is,  Xception \cite{chollet2017xception}, Convolutional Block Attention Module (CBAM) \cite{woo2018cbam} and self-attention Mechanism \cite{aug_attention_conv}.

\subsection{Xception}
\begin{figure*}[htb]
\centering
\includegraphics[scale=0.45]{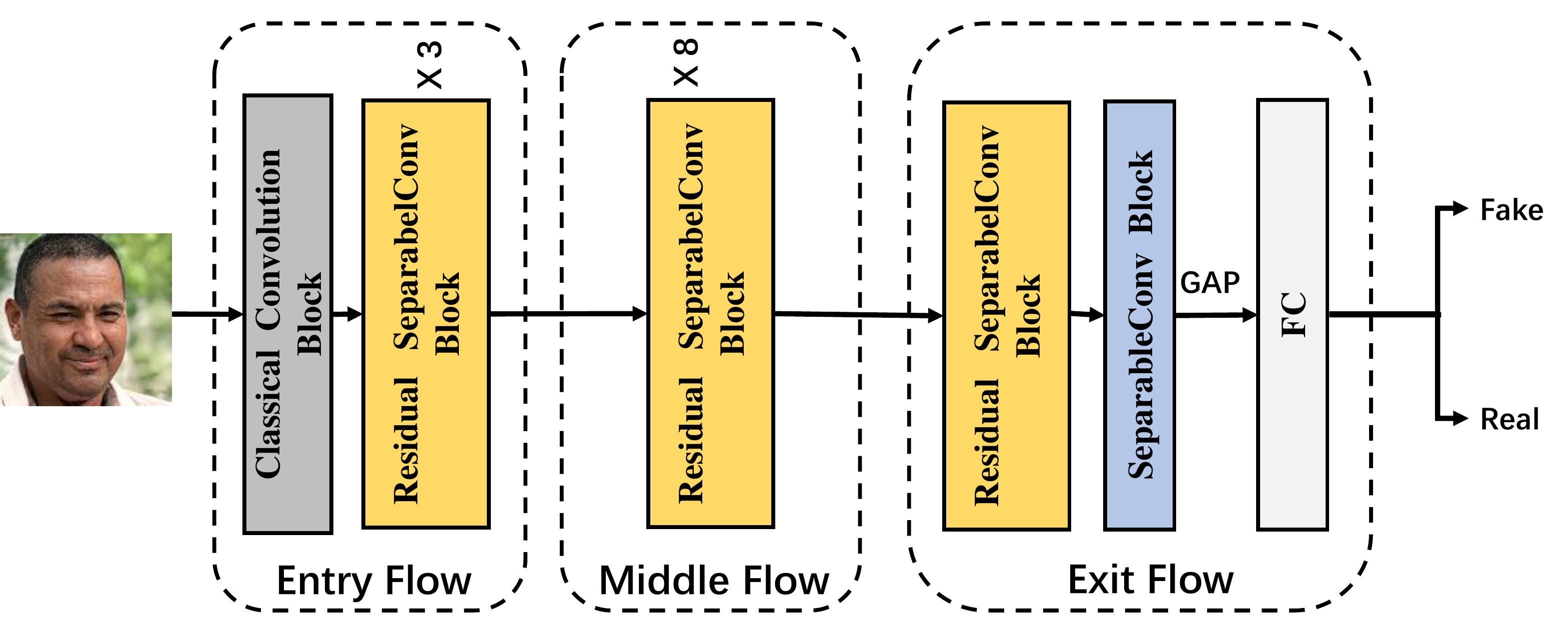}
\caption{The framework of  Xception, which includes entry flow, middle flow and exit flow.  Note that all Convolution and SeparableConvolution layers are followed by batch normalization \cite{BN} }
\label{Fig:xception}
\end{figure*}

As illustrated in Fig. \ref{Fig:xception},  Xception \cite{chollet2017xception} consists of three flows:  Entry flow,  Middle flow, and Exit flow.  Entry flow includes a  classical convolution block and 3 residual separableConv blocks;  Middle flow includes 8  residual separableConv blocks; Exit flow includes 1  residual separableConv block and 1 separableConv block.  Expect for the first block and the last block, all blocks have linear residual connections around them, which aims to prevent the gradient from disappearing during the training process of the network.  Xception is a network architecture based on depthwise separable convolution, which can not only significantly reduce the number of parameters, but also independently learn channel correlation and spatial correlation separately.  Note that Xception is originally used for traditional image classification.  In \cite{rssler2019faceforensics},  Xception is introduced to  detect face forgery and achieves good detection results.  Thus,  Xception usually serves as a baseline network for comparative study in most recent related works. 

\subsection{Convolutional Block Attention Module}
\begin{figure}[t]
\centering
\includegraphics[width=0.5\textwidth]{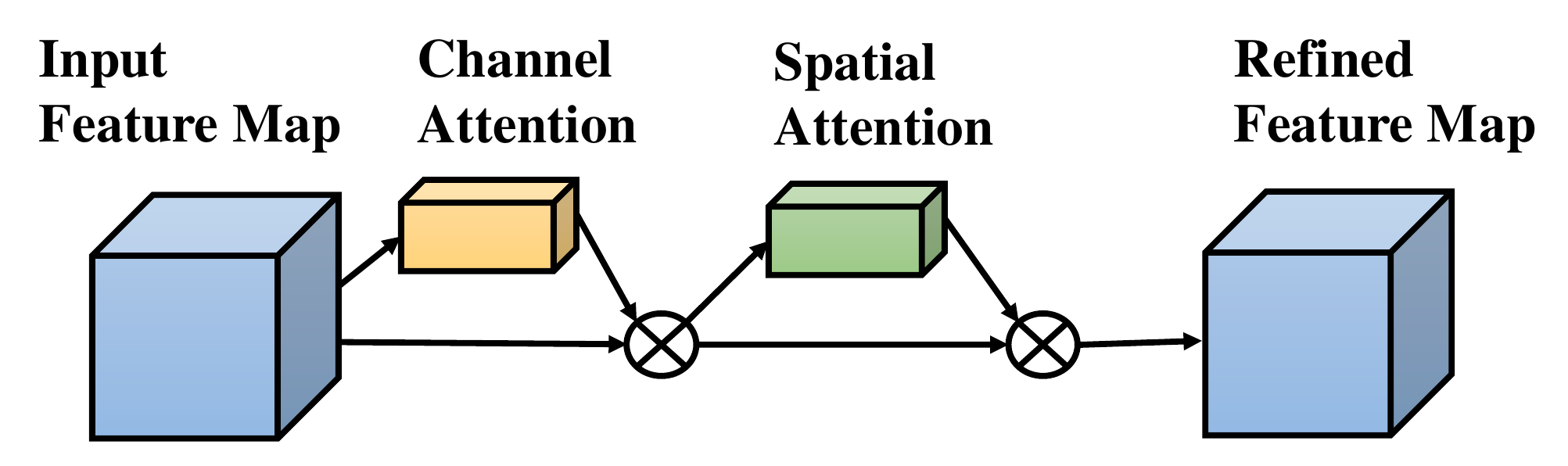}
\caption{The framework of the Convolutional Block Attention Module}
\label{Fig:CBAM}
\end{figure}

CBAM \cite{woo2018cbam} is a lightweight and general module, which aims to refine features via attention mechanism.   As illustrated in  Fig \ref{Fig:CBAM},  given an input feature map,  CBAM   infers  attention maps along two separate dimensions, i.e.,  channel and spatial, then the attention maps are multiplied to the input feature map for adaptive feature refinement.  In this way,  CBAM can help the model refines the feature effectively by learning what and where to emphasize or suppress.  CBAM is  used in sign language recognition \cite{CBAM-resnet3d}, generative model \cite{CBAM-GAN} and target detection \cite{CBAM-target-detection}.   To our best knowledge, however, there are no related works for deepfake detection.

\subsection{Self-Attention Mechanism}
\label{subsec:Self-Attention}
 The self-attention mechanism \cite{self_attention} tries to focus on key information and ignore irrelevant information via  learning different weights corresponding to the feature maps.  The self-attention mechanism can be described as mapping a query and a set of key-value pairs to an output, where the query (Q), keys (K) and values (V) can be obtained from the input feature map.  
Let  $H$, $W$, $F_{in}$ denote the height, width and number of channel of input feature map. $d_{k}$ denotes the keys of dimension.
Given an input feature map of shape $( H,W,F_{in})$, we firstly flatten it to a matirx $X\in$ $\mathbb{R}^{HW\times F_{in}}$. The output of the single-head attention mechanism can be calculated as:\\
\begin{equation}
\label{formula_self_attention}
\begin{aligned}
Q = XW_{q}; K = XW_{k};  V = XW_{v}\\
output  = softmax( \frac{ Q K^{T}}{\sqrt{d_{k}  } } ) V
\end{aligned}
\end{equation}

\noindent where $W_{k}, W_{q}$ and $W_{v}$  represent randomly initialized coefficient matrices,  which can be updated  during training stage. The softmax denotes the normalization operation.  Through the self-attention mechanism,  we can adaptively learn the global feature information. Note that the classical convolution operation  has a significant weakness in that it only operates on a local neighbourhood,  thus missing global information.  In some applications,  self-attention mechanism can be  used to  enhance  the convolution operation.

\section{Proposed Method}
\label{sec:Proposed}
\begin{figure*}[t]
\centering
\includegraphics[scale=0.56]{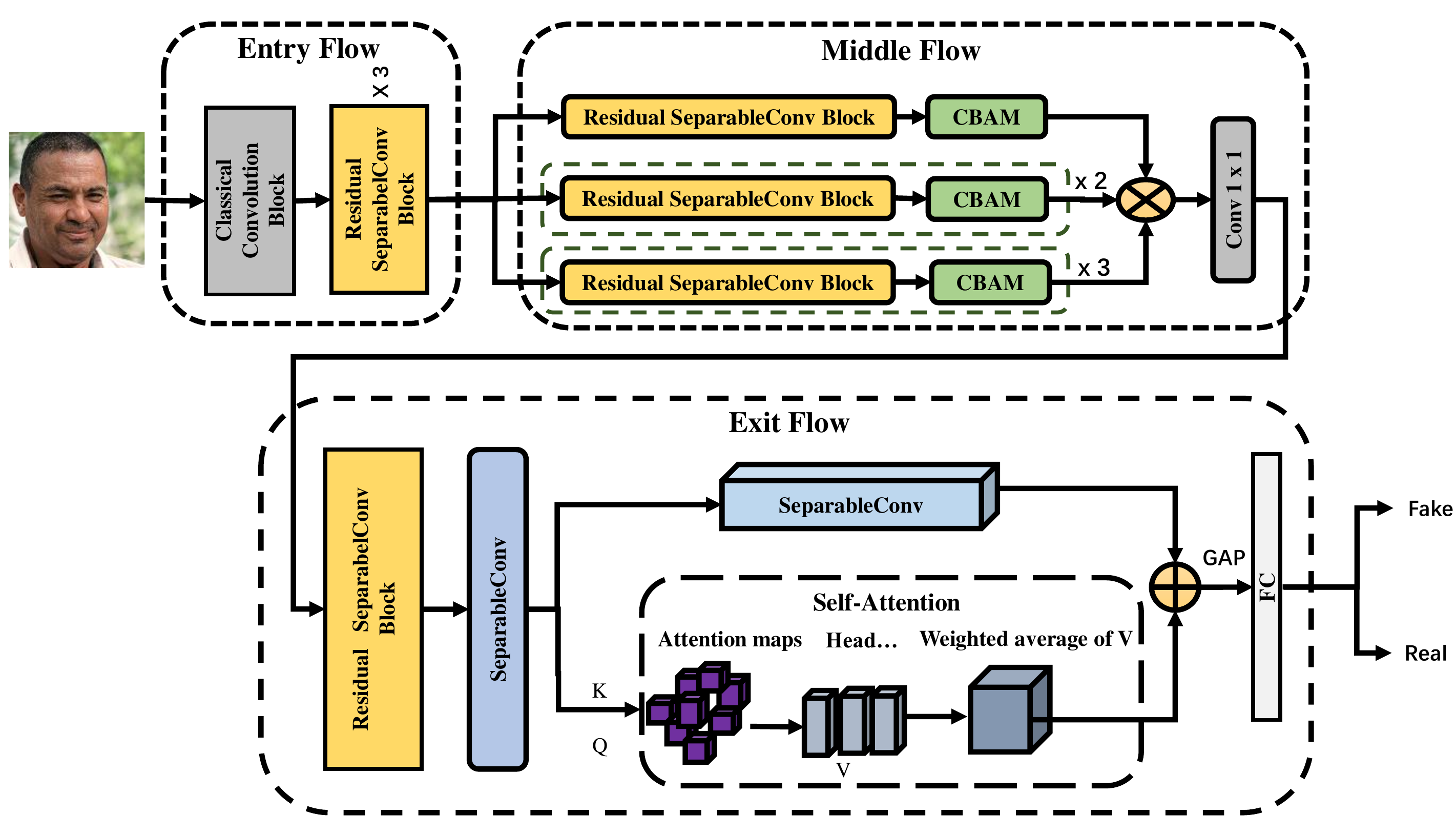}
\caption{The proposed framework based on Xception. The symbol $\otimes$ denotes the feature map concatenation,  and $\oplus$ denotes feature map addition. }
\label{fig:model}
\end{figure*}

As illustrated in Fig. \ref{fig:model},  the proposed method is an improved version of  Xception,  and it also includes three flows, that is, entry flow,  middle flow and exit flow.  We will describe the three modules in the following subsections.

\subsection{Entry Flow}
The entry flow in the proposed framework is exactly the same as that in original Xception model (refer to Fig. \ref{Fig:xception}), which consists of a classical convolution block and 3 residual separabelconv blocks.   

\subsection{Middle Flow Combined with Feature Fusion \& CBAM}
The middle flow in  the original Xception is composed of 8 residual separableConv blocks within a single branch.  In the proposed middle flow, however, we firstly reduce the number of residual separableConv blocks from 8 to  6,  and  assign them into three parallel branches.  As illustrated in Fig. \ref{fig:model},  we assign 1, 2, and 3  residual separableConv blocks in the three branches respectively.  In this way,  the proposed method is able to learn different  high-dimensional semantic features of  image  by using different levels of convolution.  Furthermore, we introduce the CBAM  after every residual separabelConv block  to enhance the characterization ability of the features and refines intermediate features.  Finally,  we  merge the reorganized high-dimensional features in the three branches, and employ a $1 \times 1$ convolution to learn the channel correlation to fully integrate these features.   By doing so,  we can also significantly  reduce the  dimension of feature map that feed to the exit flow.

\subsection{Exit Flow Combined with Self Attention}

The exit flow in Xception consists of a residual separableConv block and a separableConv block (including two depthwise separable convolutions). Thus, just local information of the fused features is considered here.  Inspired by method \cite{aug_attention_conv},  we introduce a self-attention  module after the first depthwise separable convolution, and this module  is paralleled with another depthwise separable convolution in the proposed exit flow as illustrated in Fig.\ref{fig:model}.  As described in section \ref{subsec:Self-Attention},  each attention map over the fused features are firstly computed from keys (K) and query (Q) in self-attention mechanism. These attention maps are used to compute weighted averages of the values (V). Then we concatenate the results and reshape them to match the original spatial dimensions via a pointwise convolution.  In addition, we employ the depthwise separable convolution at the other branch to learn the local information of the fused features.  Finally, we combine  the global information (via self-attention)  and local information (via depthwise separable convolution) of the fused features, and feed them to  a fully connected layer for classification.

\section{Experiments}
\label{sec:experiments}

\begin{table*}[htb]
\renewcommand\arraystretch{1.5}
\caption{Detection accuracies of different detection networks.  Some results are obtained from \cite{zi2021wilddeepfake}.  In all following Tables, those values with an asterisk (*)  denote the best  performances in the corresponding cases. }
\centering
\setlength{\tabcolsep}{7mm}{
\begin{tabular}{cccc}
\hline
Network   &  TIMIT (LQ) & FF++deepfake  (LQ) & WildDeepfake \\
\hline
AlexNet \cite{AlexNet}         & 94.77\%  & 90.58\% & 60.37\% \\
VGG16 \cite{vgg16}          & 98.73\%  & 90.19\% & 60.92\% \\
ResNetV2-50 \cite{resNet}   & 94.88\%  & 90.91\% & 63.99\% \\
ResNetV2-152 \cite{resNet}     & 95.68\%  & 88.00\% & 59.33\% \\
Inception-v2 \cite{inceptionV3}    & 90.30\%  & 89.44\% & 62.12\% \\
MesoNet-4 \cite{mesoNet}        & 91.18\%  & 87.75\% & 64.47\% \\
MesoNet-Inception\cite{mesoNet}         & 97.85\%  & 84.82\% & 68.48\% \\
Xception \cite{rssler2019faceforensics}        & 99.65\%  & 90.25\% & 74.32\% \\
ADDNet-2D \cite{zi2021wilddeepfake}         & 99.54\%  & 90.42\% & 76.25\% \\
Capsule \cite{capsule}         & 99.85\%  & 93.81\% & 77.60\% \\
Proposed Method        & \bf{99.86\%*}  & \bf{96.24\%*} & \bf{83.32\%*} \\
\hline
\end{tabular}}
\label{Table:Comparative}
\end{table*}

In our experiments, we employ three popular databases, including Deepfake-TIMIT ( denoted as TIMIT for short) \cite{korshunov2018deepfakes}, FaceForensics++  (denoted as FF++)   \cite{rssler2019faceforensics}, and WildDeepfake \cite{zi2021wilddeepfake}. In training stage,  we use MTCNN \cite{zhang2016joint}  to locate and align the face regions in video frames, and then resize the face images to the size of $224\times224$.  For a fair comparison,  all detection results are evaluated on the same test dataset as existing methods. 

We employ the cross-entropy loss and Adam optimizer with the batch size 64.  We set the  initial learning rate to be  0.0001.  We  train the networks in total of 20 epochs,  and reduce the learning rate by a half  every 5 epochs, and select  the model with the best accuracy as the final model.  

\subsection{Evaluation on FF++, TIMIT and WildDeepfake}

In this experiment, just the low-quality (LQ) versions of TIMIT and FF++(deepfake) databases are considered  since  their high-quality (HQ) versions are  relatively detected by  the current methods.   The detection accuracies are  shown in Table \ref{Table:Comparative}.  From Table  \ref{Table:Comparative}, we obtain two following observations:
\begin{itemize}
\item First of all, the proposed method always outperforms other test methods in all cases.  Compared with the advanced method (i.e., Capsule \cite{capsule}),  the proposed method achieves 0.01\%, 2.43\%, and 5.72\%  improvements for TIMIT (LQ), FF++deepfake (LQ), and WildDeepfake  respectively.  Compared with the original  Xception \cite{rssler2019faceforensics} model,  the proposed method  achieves 0.21\%, 5.99\%, and 9.00\%   improvements for the three databases respectively, which means that the feature fusion and  dual attention mechanism introduced in the proposed model can effectively enhance the ability of the  original Xception for face forgery detection.

\item Compared with TIMIT and FF++,  the detection accuracies for WildDeepfake are relatively lower for all detection networks.  The main reason is due to the diversity of WildDeepfake.  Note that those video clips of  WildDeepfake are collected on the Internet, they  are from different sources and may be compressed several times. Based on our experiments,   the proposed method still obtain 83.32\% detection accuracy for WildDeepfake. While the detection accuracies are all lower than 77.70\% for other test networks. 

\end{itemize}

\subsection{Cross-dataset Evaluation on Celeb-DF}
To show the generalization of the proposed method,  cross-dataset evaluation on a new database - Celeb-DF \cite{Celeb_df} is considered.  In this section,   we first  train a  classifier  based on  the high-quality version of FF++ and WildDeeepfake respectively, and then use the resulting classifier to detect  those face forgeries on Celeb-DF.   The comparative results  are shown in Table \ref{Table:Cross1} and Table \ref{Table:Cross2} respectively.

\begin{table}[t]
\renewcommand\arraystretch{1.5}
\center
\caption{Cross-dataset evaluation on Celeb-DF (AUC) via training on FF++(HQ).  Some results  are  obtained from \cite{zhao2021multiattentional}.}
\begin{tabular}{ccc}
\hline
network         & FF++    & Celeb-DF   \\
\hline
MesoNet-Inception \cite{mesoNet}  & 83.00\%    & 53.60\% \\
Xception\cite{rssler2019faceforensics}    & {99.70}\% & 65.30\% \\
Capsule \cite{capsule}    & 90.61\% & 67.92\% \\
Proposed Method            & 98.09\%     & \bf{68.39\%*} \\
\hline
\end{tabular}
\label{Table:Cross1}
\end{table}

\begin{table}[t]
\renewcommand\arraystretch{1.5}
\center
\caption{Cross-dataset evaluation on Celeb-DF(ACC) via training on WildDeepfake. }
\begin{tabular}{ccc}
\hline
network         & WildDeepfake    & Celeb-DF   \\
\hline
MesoNet-Inception \cite{mesoNet}  & 68.48\%    & 49.11\% \\
Xception\cite{rssler2019faceforensics}    & 74.32\% & 51.87\% \\
Capsule \cite{capsule}    & 77.60\% & 53.00\% \\
Proposed Method            &{83.32}\%     & \bf{72.62\%*} \\
\hline
\end{tabular}
\label{Table:Cross2}
\end{table}

The results in the two above tables show that  the generalization of the proposed method is  better than  three other modern networks. Especially for the case of training on Wilddeepfake, the proposed method achieves around 20\% improvements.

\subsection{Ablation Study}
\label{ablation}
In the proposed method, we introduce  dual attention mechanism (i.e., CBAM in middle flow and self-attention in exit flow) and feature fusion (in the middle flow) into original Xception model  for face forgery detection.  In order to verify the rationality  of  the proposed method,  we conduct the several ablation experiments in this section.  For simplicity,  WildDeepfake  is used for performance evaluation.

\begin{table}[t]
\caption{Ablation results on WildDeepfake(ACC)with different attention mechannisms. }
\center
\renewcommand\arraystretch{1.5}
\setlength{\tabcolsep}{2mm}{
\begin{tabular}{cc}
\hline
Setups                     & WildDeepfake \\
\hline
Proposed Method w/o CBAM\&Self-Attention                  & 78.28\%      \\
Proposed Method w/o Self-Attention           & 80.09\%      \\
Proposed Method w/o CBAM & 82.78\%      \\
Proposed Method                   & \textbf{83.32\%*}      \\
\hline
\end{tabular}}
\label{ablation_study}
\end{table}

\begin{itemize}
\item Analysis on Dual Attention Mechanism:  In this experiment, we evaluate the the proposed model after removing the CBAM and/or self-attention. The experimental results are shown in Table \ref{ablation_study}.  From Table \ref{ablation_study},  we observe that when both  CBAM and self-attention are  removed from the proposed model,  the detection accuracy is  78.28\%.  When  self-attention  or CBAM  is removed,  the corresponding detection accuracies increase to 80.09\% and 82.78\% respectively.  When  the self-attention  and  CBAM are preserved, the proposed method works the best, and achieves 83.32\% detection accuracy. The above results show that the dual attention mechanism is useful for face forgery detection.  

\vspace{0.5em}

\item Analysis on Different Setups in Middle Flow:  In the proposed middle flow,  we design three branches (i.e., 1, 2, and 3 residual SparavleConv Blocks and CBAM respectively) to extract different high-dimensional semantic features of face image, and then perform feature fusion on all branches for subsequent network analysis.  In this experiment,  we evaluate the performance on using different setups in the middle flow.  Some comparative results are shown in Table \ref{tab:middle_flow}.  From Table \ref{tab:middle_flow}, we observe that  the proposed middle flow achieves the best detection performance.  When adding the 4th branch  (i.e.,  4 residual SparavleConv Blockss and CBAM)  or removing the 3rd branch in proposed middle flow,  the corresponding detection accuracies will drop from 83.32\%  to 79.20\% and 74.99\% respectively, which means that setting 3 branches is reasonable  in the proposed model.  In addition, we compared  other four setups with a single branch, that is,  just using  the 1st, 2nd, 3rd branch respectively,  and the original middle flow in Xception.  Their detection accuracies  all are lower than  80.05\%,  which means that feature fusion   is helpful to enhance the model performance. 

\end{itemize}

\section{Conclusion}
\label{sec:Conclusion}
In this paper,  we first introduce dual attention mechanism (i.e., CBAM and self-attention) and feature fusion into  Xception for face forgery detection, and show that the proposed method can significantly improve the detection performance of  the original  Xception model,  and achieve state-of-the-art results both in the effectiveness and generalization ability compared with related methods.   In addition, we also provide some ablation experiments to verify the rationality of introducing dual attention mechanism and feature fusion in the  proposed method.   In future, there are several important issues in the proposed framework worthy of in-depth study.  For instance,  we just consider the spatial information as network input in the proposed method.  We would  combine some features in the frequency domain, and construst  a two  stream  network for further enhancing the detection performance.  
In addition to high-dimensional facial semantic features, we try to use multiple attention maps to explore discriminative local region (such as eyes and nose) on the face,  and  fuse these features  in the middle flow for face forgery detection.

\begin{table}[t]
\caption{Ablation results on WildDeepfake (ACC) with different  setups in the proposed middle flow.   }
\center
\renewcommand\arraystretch{1.5}
\setlength{\tabcolsep}{1mm}{
\begin{tabular}{c c c}
\hline
  Setups in Middle Flow         &   Branch Number &   WildDeepfake \\
\hline
Proposed middle flow       &  3      &  \bf{83.32\%*}     \\
Adding the 4th branch       & 4 & 79.20\%      \\
Removing the 3rd branch  & 2 & 74.99\%     \\
Just using the 1st branch    & 1 &   79.60\%     \\
Just using the 2nd branch   & 1 &  79.57\%     \\
Just using the 3rd branch   & 1 &   80.04\%     \\
Original middle flow in Xception    & 1  &  78.53\%     \\
\hline
\end{tabular}}
\label{tab:middle_flow}
\end{table}

\bibliographystyle{IEEEtran}
\bibliography{cited}

\end{document}